\documentclass[10pt,twocolumn,letterpaper]{article}

\usepackage{cvpr}
\usepackage{times}
\usepackage{epsfig}
\usepackage{graphicx}
\usepackage{amsmath}
\usepackage{amssymb}
\usepackage{multirow}
\usepackage{cite}
\usepackage{subfig}
\usepackage{enumerate}
\usepackage{authblk}
\usepackage{algorithm}
\usepackage{algorithmic}

\usepackage{hyperref}

\cvprfinalcopy 

\def\cvprPaperID{5137} 

\def\etal{\emph{et al.}}

\ifcvprfinal\fi
\begin{document}

\title{\vspace{-2mm}Context-aware and Scale-insensitive Temporal Repetition Counting\vspace{-9mm}}
\author[1]{Huaidong Zhang}
\author[1,2,3]{Xuemiao Xu ${}^*$}
\author[1]{Guoqiang Han}
\author[1]{Shengfeng He\thanks{Xuemiao Xu and Shengfeng He are joint corresponding authors. Email: xuemx@scut.edu.cn, hesfe@scut.edu.cn}
\vspace{-2mm}}
\affil[1]{South China University of Technology}
\affil[2]{State Key Laboratory of Subtropical Building Science}
\affil[3]{Guangdong Provincial Key Lab of Computational Intelligence and Cyberspace Information}

\teaser{\vspace{-10mm}
\centering
\subfloat{\includegraphics[width=1.0\linewidth]{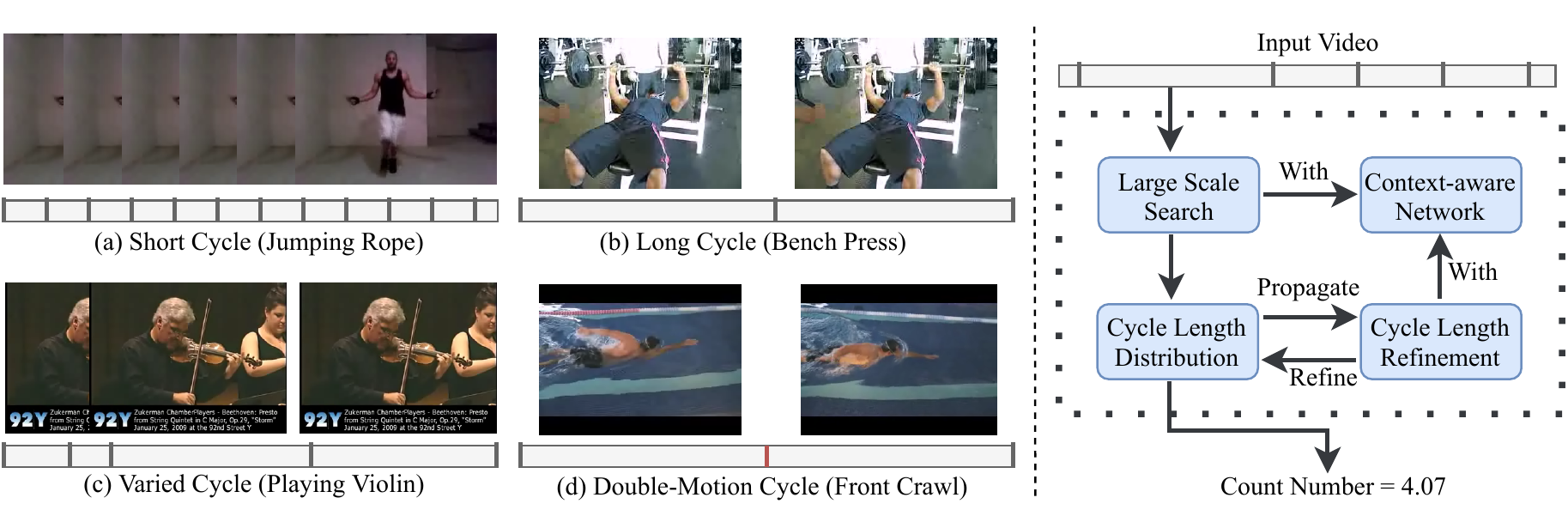}}
\vspace{-3mm}\caption{The varied cycle lengths and context-dependent motions of repetitive actions pose challenges for counting temporal repetitions. We propose a context-aware and scale-insensitive framework to cope with these problems. The counting process is designed in a coarse-to-fine manner, integrating with a context-aware network for detecting bidirectional repetitive actions.}\label{fig:teaser}\vspace{-12mm}
}

\maketitle
\thispagestyle{empty}

\begin{abstract}\vspace{-5mm}
Temporal repetition counting aims to estimate the number of cycles of a given repetitive action. Existing deep learning methods assume repetitive actions are performed in a fixed time-scale, which is invalid for the complex repetitive actions in real life. In this paper, we tailor a context-aware and scale-insensitive framework, to tackle the challenges in repetition counting caused by the unknown and diverse cycle-lengths. Our approach combines two key insights: (1) Cycle lengths from different actions are unpredictable that require large-scale searching, but, once a coarse cycle length is determined, the variety between repetitions can be overcome by regression. (2) Determining the cycle length cannot only rely on a short fragment of video but a contextual understanding. The first point is implemented by a coarse-to-fine cycle refinement method. It avoids the heavy computation of exhaustively searching all the cycle lengths in the video, and, instead, it propagates the coarse prediction for further refinement in a hierarchical manner. We secondly propose a bidirectional cycle length estimation method for a context-aware prediction. It is a regression network that takes two consecutive coarse cycles as input, and predicts the locations of the previous and next repetitive cycles. To benefit the training and evaluation of temporal repetition counting area, we construct a new and largest benchmark, which contains 526 videos with diverse repetitive actions. Extensive experiments show that the proposed network trained on a single dataset outperforms state-of-the-art methods on several benchmarks, indicating that the proposed framework is general enough to capture repetition patterns across domains. Code and data are available in \url{https://github.com/Xiaodomgdomg/Deep-Temporal-Repetition-Counting}.
\end{abstract}

\vspace{-3mm} \section{Introduction}
\label{sec::introduction}
Human activities are commonly involved repetitive actions. Temporal repetition counting is a problem that aims to count the number of repetitive actions in a video~\cite{pami00_cutler,cvpr08_pogalin,iccv15_levy,ijcv19_runia}. The repetition analysis is explored as an auxiliary cue to other video analysis applications, such as cardiac and respiratory signal recover~\cite{li2018repetitive}, pedestrian detection~\cite{ran2007pedestrian}, 3D reconstruction~\cite{li2018structure,ribnick20103d}, and camera calibration~\cite{huang2016camera}.

This is a challenging problem as repetitive actions exhibit inherently different action patterns. We summarize 4 representative cases in the left part of Figure~\ref{fig:teaser}. Figure~\ref{fig:teaser}(a) and (b) show the most common repetitions, in which actions are performed in fixed cycles. The problem of detecting these two repetitions is that their cycle lengths varied largely, and therefore is invalid to make restricted assumptions about the time-scale of the cycle length across actions. In Figure~\ref{fig:teaser}(c), the case of playing the violin shows that the cycle lengths are not always a fixed value. This case is contradictory to (a) and (b), and hence the assumption of actions will be performed in a periodic manner is false. In Figure~\ref{fig:teaser}(d), a front crawl action can be decomposed into two sub-actions with a similar motion field, crawling with the left hand and right hand.
As the two sub-actions are similar in motion space, contextual information in semantic space should be considered to avoid the double counting error.

Most existing methods ~\cite{cvpr08_pogalin,pami07_briassouli,pami00_cutler,iccv15_levy,li2018repetitive} rely heavily on the periodicity assumption. As a consequence, although the representative work~\cite{iccv15_levy} achieves a near-perfect performance on the periodic dataset \emph{YTsegments}, it cannot detect varied cycle lengths in the non-stationary video dataset \emph{QUVA Repetition}~\cite{cvpr18_runia}. While the latest work~\cite{ijcv19_runia} address this problem, it detect repetition solely based on the motion field. Therefore it conflicts with the scenarios like Figure~\ref{fig:teaser}(d), in which repetitions cannot be distinguished by motion field and contextual and semantic information is required to understand the action. Based on the above observations, we argue that detecting repetitions should 1) exhaustive search for a large range of cycle lengths to cover most unknown actions; 2) include contextual understanding and estimating cycle lengths by taking multiple periods into consideration.

In this paper, we tailor a context-aware and scale-insensitive framework based on the above principles. The data flow is shown in the right part of Figure~\ref{fig:teaser}. Following rule~\#1 to exhaustively search all the time scales can absolutely address the cycle lengths variations problem, but it leads to expensive computation. We combat this problem by proposing a coarse-to-fine cycle lengths estimation strategy integrated with a regression network. In particular, we only exhaustive search the initial cycle lengths for a local video clip. The initial estimation, is then propagated to the entire video, and each of the estimated repetition in the video is refined by our regression model. In this way, we largely reduce the computational cost in searching accurate cycle lengths, while we can adapt to large variations of cycle lengths in the same video. The proposed regression model handles rule~\#2, in which we inject contextual information for estimating accurate cycle lengths. Specifically, instead of taking only one action cycle as input, we sample the video to contain two consecutive repetitions, named double-cycle. Given such broad context, our regression model aims to relocate the previous and future repetitive cycles in a bidirectional manner. Furthermore, existing researches in repetition counting lack of sufficient data, therefore we propose a new repetitive action counting benchmark, named \emph{UCFRep}. It is constructed by annotating repetitive actions from the widely used dataset \emph{UCF101}~\cite{soomro2012ucf101}, and it is the largest dataset containing 526 videos. Extensive experiments demonstrate the proposed method is able to cope with various repetitive actions, and we outperform state-of-the-art methods on three benchmarks.

Our contributions are four-fold:
\begin{enumerate}[(1)]
\vspace{-2mm}   \item We propose a coarse-to-fine double-cycle estimation strategy integrated with regression, which allows fast estimation of cycle lengths for the entire video and dynamic relocation of varied cycles.
\vspace{-2mm}   \item We present a bidirectional context-aware regression model. It explores contextual information to simultaneously estimate the previous and future cycles in a bidirectional manner.
\vspace{-2mm}   \item We construct a new and largest benchmark \emph{UCFRep}. 526 repetitive action videos are annotated for training and evaluation.
\vspace{-2mm}   \item The proposed network outperforms state-of-the-art methods on three benchmarks, especially we achieve superior performances on two unseen benchmarks (without fine-tuning). It reveals the proposed framework is general enough to complex and unknown scenes.
\end{enumerate}

\section{Related Work}
\label{sec:related_work}
A typical solution for temporal repetition counting is to transfer the motion field into one-dimensional signals, and then they try to recover the repetition structure from the signal period~\cite{iccv05_laptev,icip18_panagiotakis,siam18_tralie,pami04_lu,albu2008generic}. The mainstream of these methods obtains repetition frequency with Fourier analysis~\cite{Azy2008segmentation,pami00_cutler,cvpr08_pogalin,pami07_briassouli}. In addition, they detect the cycle by filtering~\cite{tip06_burghouts}, peak detection~\cite{thangali2005periodic}, classification~\cite{davis2000categorical}, and singular value decomposition~\cite{chetverikov2006motion}. The above methods assume that the estimating repetition is periodic, so that they cannot handle the non-stationary repetitions.
A recent work~\cite{ijcv19_runia} addresses this limitation, and propose a novel inference scheme to detect non-stationary actions.
However, they only adopt the motion field to extract features for analysis, while ignoring context-dependency in semantic domain.

Like us, there are methods that also use deep features for repetition analysis. Li~\etal~\cite{li2018repetitive} propose to learn temporal dependency by adopting the LSTM network on the sequence of images. They aim to recover the cardiac and respiratory signals from the medical image sequence, as such their method cannot handle complex repetitions in real-world. Levy and Wolf~\cite{iccv15_levy} aim to propose a classification network for live repetition estimation. Their network is designed to extract features of 20 frames from the video with the predefined sampling-rate. As discussed above, a predefined cycle lengths cannot adapt to the complex repetitive actions with large variations of cycle lengths.

\begin{figure*}
    \centering
    {\includegraphics[width=1.0\linewidth]{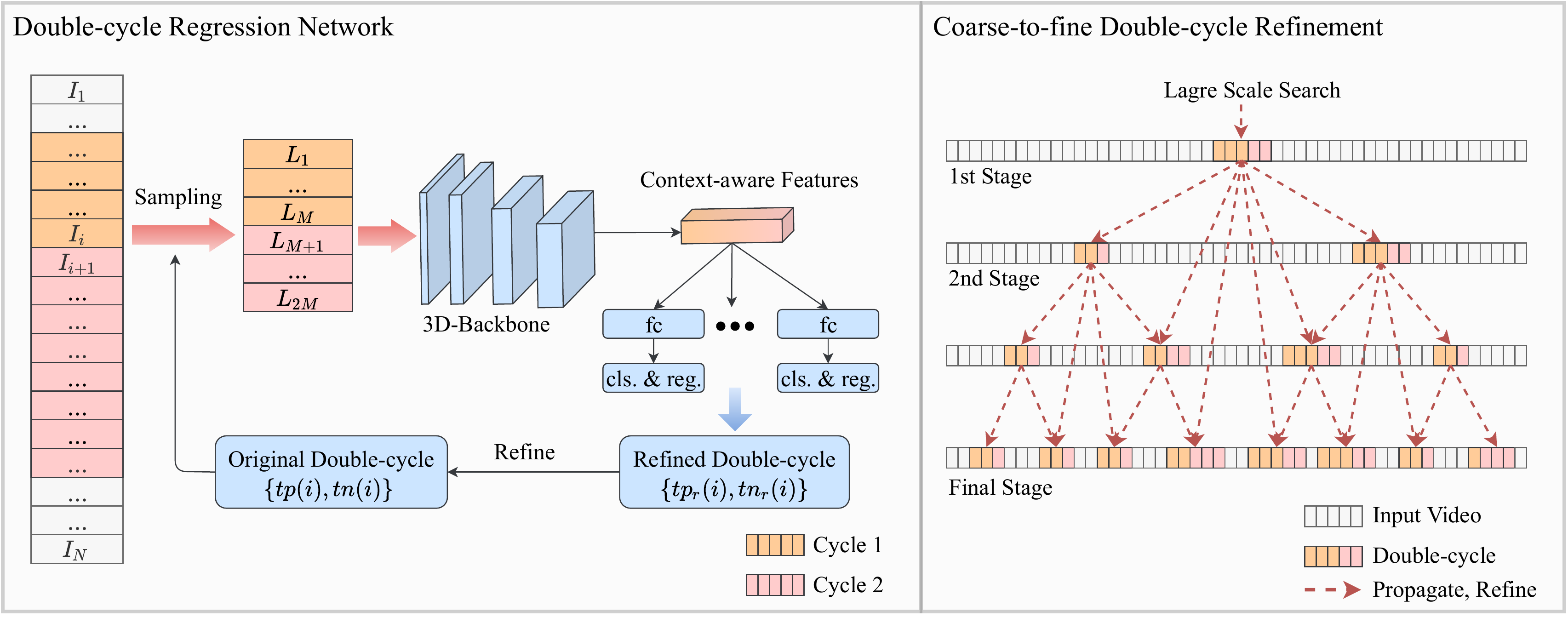}}
    \vspace{-8mm}\caption{Framework overview. The proposed context-aware double-cycle regression network is shown on the left. It regresses a new double-cycle $\{{tp}_r(i),{tn}_r(i)\}$ based on the context information sampled from the previous double-cycle $\{{tp}(i),{tn}(i)\}$. In the right part, a coarse-to-fine double-cycle refinement method is illustrated. We first perform exhaustive search locally on the first stage, and the initial double-cycle is propagated and refined in the following stages. An accurate counting result can be obtained by averaging all the cycle lengths in the video.}\label{fig:overview}\vspace{-5mm}
\end{figure*}

Action localization~\cite{lin2018bsn,shou2016temporal,Long_2019_CVPR} shares a similar spirit to localize actions in temporal domain. These methods aim to locate the temporal begin and ending points of each action in the entire video, hence these methods can be easily adapted to the field of repetition counting.
However, these methods find the action segment separately, which means that they ignore the repetition priors to effectively utilize the context information.
In our method, we borrow the idea of the anchor-based temporal regression from this literature, and further explore context dependency.

\section{Approach}
\label{sec:method}
In this section, we first introduce the problem formulation and overview of the proposed context-aware and scale-insensitive framework. Then we describe two core modules of our framework, the context-aware double-cycle regression network and coarse-to-fine double-cycle refinement. Finally, we present the details of our newly constructed temporal repetition benchmark.

\subsection{Problem Formulation}
\textbf{Repetition definition.} We have a different problem setting than prior works, as we aim to locate both previous and future cycles in a bidirectional manner. Given a video with $N$ frames $\mathcal{I}=\{I_1,I_2,...,I_N\}$, the repetition can be defined as follows: for a frame $I_i$, if we can find a previous frame $I_{p(i)}$ and a future frame $I_{n(i)}$, such that the two frame sequences $\{I_{p(i)}, I_{p(i)+1},...,I_{i}\}$ and $\{I_i, I_{i+1},...,I_{n(i)}\}$ contains the identical actions, then there are two repetitions existing in these two sequences. We refer these two consecutive cycles as \emph{double-cycle}, and the $I_{p(i)}$ as the \emph{previous repetitive frame} of $I_i$ and $I_{n(i)}$ as the \emph{next repetitive frame} of $I_i$.

\textbf{Target formulation.} In this paper, we aim to count the temporal repetition number $c$ for the given video.
If the action is strongly periodic in the video, we can assume the cycle length is a constant across the entire video. Then we can easily estimate the repetitions number by finding the previous and next repetitive frame locations $\{p(i),n(i)\}$ of an arbitrary frame $i$ and calculate the number of repetitions $c$ as:
\begin{equation}
c=\frac{N}{i-p(i)+1}=\frac{N}{n(i)-i+1}.
\end{equation}

However, the variety between repetitions cannot be neglected in the real-world. To tackle this problem, we propose to calculate the repetition counts by estimating $\{p(i),n(i)\}$ of each frame in the video. Therefore we formulate the problem as
\begin{equation}
c = \sum_{i=1}^{N}\left (\frac{0.5}{i-{p}(i)+1}+\frac{0.5}{{n}(i)-i+1} \right ).
\label{eq:eq2}
\end{equation}
Two cycle lengths can be computed as ${tp}(i)=i-{p}(i)+1$ and ${tn}(i)={n}(i)-i+1$. For clarity, we define $\{{tp}(i),{tn}(i)\}$ as the \emph{double-cycle} that describes two consecutive repetitions with frame $i$. 

\subsection{Framework Overview}
Following the target formulation, our framework is designed to predict the double-cycle $\{{tp}(i),{tn}(i)\}$ for all the position $i\in \{1,2,...,N\}$.
We first propose a context-aware double-cycle regression network, which is illustrated in the left part of Figure~\ref{fig:overview} and described in Section~\ref{sec:model}. The network is designed to refine the given double-cycle for a specific position. Given an initial double-cycle, our network extracts the 3D features based on some sampled video frames and outputs a new double-cycle $\{{tp}_r(i),{tn}_r(i)\}$. With the extracted context information from a large range of video frames, the network is able to identify the repetition and regress the cycle lengths easily. Furthermore, this process is performed multiple times to obtain a progressively refined double-cycle.

As discussed above, an exhaustive search should be performed to cope with the large cycle length variation problem. It can also provide an reasonable initial double-cycle for the regression network. Instead of searching the entire video, we first search locally in the video, and propagated the prediction to the other frames.
The right part of Figure~\ref{fig:overview} shows our method and it is described in Section~\ref{sec:refinement}. We perform exhaustive searching for one time in the middle frame of the video, such that the initial double-cycle is likely within the same scale with others. It is then propagated to the other frames, each of the new frame is integrated with the regression network for local refinement. For each stage we sample the positions uniformly across the video so that the sampled position can be the propagation root for the next stages.
The final repetition counts of the video can be calculated by the repetition count summarization of all frames.

\subsection{Double-cycle Regression Network}
\label{sec:model}
The objective of the network is to refine the input double-cycle $\{{tp}_r(i),{tn}_r(i)\}$ of an assigned position $i$.
To extract features of fixed size for regression, we sample specific frames within the double-cycle. As illustrated in the left part of Figure~\ref{fig:overview}, network input $L$ is a sequence with $2M$ frames, which consists of two half. We sample the first half of the inputs uniformly from the range $[i-2{tp}(i),i]$, and the next half inputs from the range $[i+1,i+2{tn}(i)+1]$.
Note that we double the sampling range to detect large context like the double-motion in Figure~\ref{fig:teaser}(d).
The sampled sequence $L$ is then fed into a 3D-backbone model. We use the 3D-ResNext101~\cite{hara2018can,xie2017aggregated} pretrained on the ActivityNet~\cite{caba2015activitynet}.
Other network architectures are also applied, please refer to the experiments for details.
We remove the last classification layer and use the outputs after pooling to be the context-aware 1D-features (4096 dimensions for ResNext101).
The features are then fed into the newly added prediction branch for classification and regression.
The prediction branch is a two fully-connected layers with multi-anchor, where we use 7 anchors with default size $\{0.5,0.66,0.8,1.0,1.25,1.5,2.0\}$ to detect different size of the repetition.
Note that totally 14 anchors are used since we have two cycles $\{{tp}_r(i),{tn}_r(i)\}$.

During training, the 3D backbone and the added branch are trained end-to-end with classification loss and regression loss. With the network outputs for classification $\{y_p,y_n\}$ and for regression $\{t_p,t_n\}$, we formulate the overall loss function:
\begin{equation}
\begin{split}
\mathcal{L}=\left (\mathcal{L}_{cls}(y_p,\tilde{y_p})+\mathcal{L}_{cls}(y_n,\tilde{y_n})\right )+\\
\lambda\left (\mathcal{L}_{reg}(t_p,\tilde{t_p})+\mathcal{L}_{reg}(t_n,\tilde{t_n})\right ),
\end{split}
\end{equation}
where $\mathcal{L}_{cls}$ is the cross-entropy loss after softmax and $\mathcal{L}_{reg}$ is the smooth $L_1$ regression loss~\cite{ren2015faster}.
$\{\tilde{t_p},\tilde{t_n}\}$ is the repetition ground truth with the parameterizations of scale-invariant center translation and the log-space cycle-lengths shifting~\cite{girshick2014rich}.
$\{\tilde{y_p},\tilde{y_n}\}$ is the classification label that equals to $1$ if intersection-over-union (IoU) of double-cycle prediction and ground truth is greater than $0.5$, and $0$ otherwise.
$\lambda$ is the weighting factor that empirically set to $50$.
During inference, the objective $\{{tp}_r(i),{tn}_r(i)\}$ is equal to the regression output of the anchor which has the highest classification score.


\begin{figure*}
    \centering
    \subfloat[Cutting \protect\\   0.12-3.00 (s)]{\includegraphics[width=0.18\linewidth]{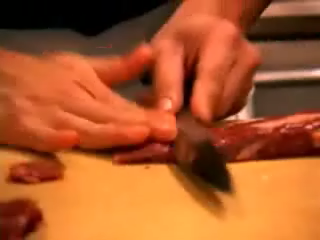}}
    \hspace{0.05cm}
    \subfloat[Hammering \protect\\ 0.24-1.88 (s)]{\includegraphics[width=0.18\linewidth]{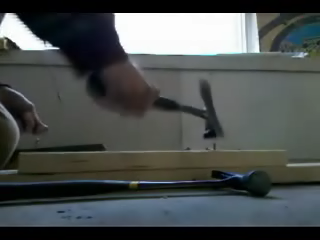}}
    \hspace{0.05cm}
    \subfloat[Shaving Beard\protect\\   0.24-3.16 (s)]{\includegraphics[width=0.18\linewidth]{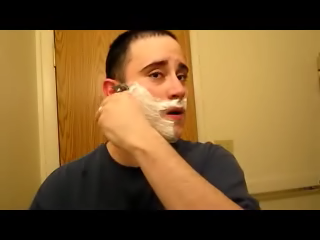}}
    \hspace{0.05cm}
    \subfloat[Hula Hoop \protect\\  0.32-0.92 (s)]{\includegraphics[width=0.18\linewidth]{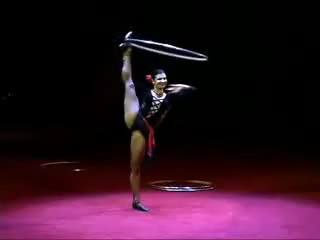}}
    \hspace{0.05cm}
    \subfloat[Soccer Juggling \protect\\   0.32-2.08 (s)]{\includegraphics[width=0.18\linewidth]{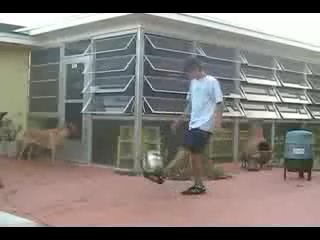}}\\
    \vspace{-0.3cm}
    \subfloat[Trampoline Jumping \protect\\ 0.60-1.56 (s)]{\includegraphics[width=0.18\linewidth]{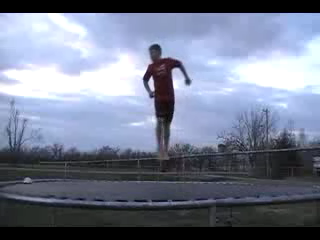}}
    \hspace{0.05cm}
    \subfloat[Biking \protect\\    0.64-2.08 (s)]{\includegraphics[width=0.18\linewidth]{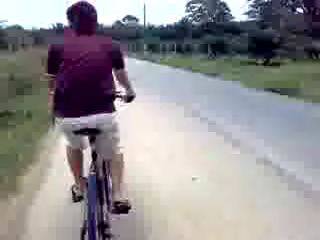}}
    \hspace{0.05cm}
    \subfloat[Table Tennis Shot \protect\\  0.64-3.20 (s)]{\includegraphics[width=0.18\linewidth]{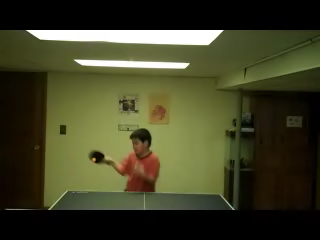}}
    \hspace{0.05cm}
    \subfloat[Hand Stand Pushups \protect\\    0.88-4.04 (s)]{\includegraphics[width=0.18\linewidth]{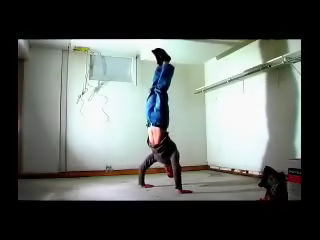}}
    \hspace{0.05cm}
    \subfloat[Rowing \protect\\    1.16-4.12 (s)]{\includegraphics[width=0.18\linewidth]{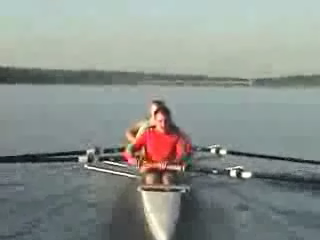}}
	\vspace{-3mm}\caption{10 examples from different categories of the \emph{UCFRep} benchmark. We annotate the minimum and maximum cycle-length of each category below the image, indicating the cycle-length variation.}\label{fig:dataset}\vspace{-4mm}
\end{figure*}

\renewcommand{\algorithmicrequire}{\textbf{Input:}}
\renewcommand{\algorithmicensure}{\textbf{Output:}}
\begin{algorithm}[t]
	\caption{Coarse-to-fine Double-cycle Refinement}
	\label{alg:Framework}
	\begin{algorithmic}[1]
    \REQUIRE Video length $N$, double-cycle regression network $F$, number of refinement stages $K$
    \ENSURE Double-cycle prediction $\{tp, tn\}$
            \STATE Initialize queue $Q$
            \STATE Determine $\{{tp}(N/2),{tn}(N/2)\}$ by global search with network $F$
            \STATE Push $\{{tp}(N/2),{tn}(N/2)\}$ into $Q$
            \FOR {\emph{k} = 1, 2, ..., K-1 }
                \STATE Initialize point set $S_k$ with $2^{k}$ points sampled uniformly over video
                \STATE Initialize $\{tp(S_k), tn(S_k)\}$ with the prediction in $Q$
                \STATE Iteratively refine $\{tp(S_k), tn(S_k)\}$ with network $F$
                \STATE Push $\{tp(S_k), tn(S_k)\}$ into $Q$
			\ENDFOR
		\RETURN $\{tp, tn\}$ in queue $Q$
	\end{algorithmic}
\end{algorithm}
\setlength{\textfloatsep}{10pt}

\subsection{Coarse-to-Fine Double-cycle Refinement}
\label{sec:refinement}
Since the network extracts the features from the context determined by the original double-cycle $\{{tp}(i),{tn}(i)\}$, a good initialization will be helpful to improve localization.
To this end, we propose a hierarchical pipeline to provide initialization by determining the double-cycle in a coarse-to-fine manner.
The key idea of the proposed pipeline is the cycle length variation between different frames can be overcome by regression, especially for the neighboring frames. Therefore each stage we refine the results on the uniformly sampled positions across the video, so that the initialization of the next stage can benefit from the neighboring prediction of the previous stage.
As illustrated in the right part of Figure~\ref{fig:overview}, in the $k$th stage, we predict $\{{tp}(i),{tn}(i)\}$ on the uniformly sampled position $i=\{N/2^k,3N/2^k,...,(2k-1)N/2^k\}$. The prediction for each position consists of two process, the initialization and refinement. Algorithm~\ref{alg:Framework} illustrates the initialization and refinement pipeline.

\textbf{Initialization.}
For the first stage, we let the double-cycle of the middle position, $\{{tp}(N/2),{tn}(N/2)\}$, equal to the value sampled from the large scale $[\mu_1,{N}/{\mu_2}]$, and then determine the initialized scale by the network classification confidence.
In the other stages, we propagate the prediction from the previous stage as initialization, following the arrow direction in the right part of Figure~\ref{fig:overview}. In particular, each position finds the previous refined neighbors for initialization. If only one neighbor is available (the first/last position of the current stage), we use it as the initialization directly. Otherwise, we merge the two observations from the previous neighbor and next neighbor averagely.
Under this scheme, we do the heavy computation search only one time in the first stage, and effectively utilize the refined results for the initialization of all the frames.

\textbf{Refinement.}
After initialization, we refine the double-cycle estimation for the given position $i$. With the refined results $\{{tp}_r(i),{tn}_r(i)\}$ from the regression network, we update the observation on position $i$ with the exponential moving average mechanism.
In other word, we update the estimation with the equation $\{{tp}(i),{tn}(i)\} = \beta\{{tp}(i),{tn}(i)\}+(1-\beta)\{{tp}_r(i),{tn}_r(i)\}$, where $\beta$ is the decay factor set as $0.5$ empirically. Note that the refinement can be performed iteratively to achieve more precise results.

After the coarse-to-fine refinement, we obtain the cycle length prediction on uniformly sampled positions.
To count the action by sampling $2^{K-1}$ points rather by all the N frames, we use the prediction of the final stage to present the prediction of all the frames by modifying Equation~\ref{eq:eq2}:
\begin{equation}
 c = \sum_{i=1}^{2^{K-1}} \frac{N}{2^{K-1}} \left (\frac{0.5}{tp(s)}+\frac{0.5}{tn(s)} \right ), s=\left \lfloor \frac{(2i-1)N}{2^K} \right \rfloor,
\end{equation}
where $K$th stage is the final stage.

\subsection{UCFRep Benchmark}
The previous repetition datasets \emph{YTsegments}~\cite{iccv15_levy} and \emph{QUVA Repetition}~\cite{cvpr18_runia} contain only 100 videos for evaluation.
Due to the lack of labeled data, the previous deep learning work~\cite{iccv15_levy} trains their model on synthesis data.
Despite the tailored design of the simulation, the domain gap between synthesis data and real data is unneglectable.
Motivated by this, we present an action repetition dataset, called \emph{UCFRep} benchmark, aiming to provide an environment for training and evaluate the data-driven model.
All the data in the proposed benchmark are collected from the widely used action recognition dataset \emph{UCF101}~\cite{soomro2012ucf101}.
Therefore, the proposed benchmark focuses on evaluating the repetition counting performance of human action.
Despite all the data is labeled with category, we find that the proposed network trained on the benchmark is general enough to perform well on the previous unseen dataset \emph{YTsegments} and \emph{QUVA Repetition} in experiments.
We mainly introduce the benchmark from three aspects, data collection, repetition labeling, and dataset statistic.

\textbf{Data collection.} The original \emph{UCF101}~\cite{soomro2012ucf101} is an action recognition data set of action videos.
13320 videos are collected from YouTube and further classified into 101 action categories.
Videos in each category are grouped into 25 groups according to whether they share common features, such as similar backgrounds, viewpoints, etc.
We check all the 101 categories from the dataset and select 23 categories in which the action is taken cyclically.
Examples of 10 categories are shown in Figure~\ref{fig:dataset}.

\textbf{Repetition labeling.} We annotate the temporal bound of repetitions similar to the principle in \emph{QUVA Repetition}~\cite{cvpr18_runia}.
Two human annotators are invited to mark out the interval contain repetitions and the repetitive frames in each video.
First, from each group in the original \emph{UCF101}, we ask the annotators to choose one video with the clearest repetitions.
If no repetitions can be founded, all the videos in this group will be abandoned.
As a result, 49 groups cannot find any repetition and $23\cdot25-49=526$ videos are collected in our benchmark.
With these videos, we let the annotators determine the repetition interval.
We consider the first frame of the interval as the reference, and ask the annotators to mark all the repetitive frames of reference within the interval.
Finally we use the average value of their annotations as the final label, and the number of repetitive frames determine the repetition counts.

\begin{table}
    \begin{center}
    \begin{tabular}{l|c|c|c}
    \hline
    {} & YTSeg & QUVA & Ours \\
    \hline
    Num. of Videos                & 100  & 100   & \textbf{526}   \\
    Duration(s)                   & 1487 & 1754  & \textbf{3500}  \\
    Num. of Counts                & 1080 & 1246  & \textbf{3506} \\
    Count Min/Max                 & 4/51 & 4/63  & 3/54  \\
    Min of Cycle(s)        & -    & 0.20  & 0.12  \\
    Max of Cycle(s)        & -    & 7.69  & 6.76  \\
    Max/Min of Cycle        & -    & 38.76 & \textbf{56.33} \\
    Cycle Variation        & 0.22 & 0.36  & \textbf{0.42}  \\
    \hline
    \end{tabular}
    \end{center}
    \vspace{-6mm}\caption{Dataset statistic of \emph{YTsegments}~\cite{iccv15_levy}, \emph{QUVA Repetition}~\cite{cvpr18_runia} and the proposed \emph{UCFRep}. Our dataset is larger than the previous datasets in terms of the number of videos, total duration and number of annotations. The wide range of cycle length between videos and large variation within the video also indicate that our benchmark is more challenges. The cycle variation is the average value of the absolute difference between minimum and maximum cycle length divided by the average cycle length.}\vspace{-2mm}\label{table:dataset}
\end{table}

\textbf{Dataset statistic.} We summarize the dataset statistic in Table~\ref{table:dataset}.
In the proposed benchmark, we provide totally 526 videos containing 3500 seconds. 3506 cycle bounds are annotated in our benchmark to provide abundant data for training and evaluation.
The benchmark also has a larger variation compared with the previous datasets.
The Max/min of Cycle indicates the difficulty from the diverse time-scale between different types of the repetitions, and the cycle variation shows the cycle-length variation within the video.

{
\begin{table*}
    \begin{center}
    \begin{tabular}{lcccccc}
    \hline
    \multirow{2}*{Method}  & \multicolumn{2}{c}{\emph{QUVA Repetition}~\cite{cvpr18_runia}}   & \multicolumn{2}{c}{\emph{YTsegments}~\cite{iccv15_levy}} & \multicolumn{2}{c}{\emph{UCFRep} (Ours)}\\
    \cline{2-7}
     &  MAE$\downarrow$ & OBOA$\uparrow$ & MAE$\downarrow$ & OBOA$\uparrow$  & MAE$\downarrow$ & OBOA$\uparrow$  \\
    \hline\hline
    Pogalin~\etal~\cite{cvpr08_pogalin} & 0.385 $\pm$ 0.376 & 0.49 & 0.219 $\pm$ 0.301 & 0.68 & - & - \\
    Levy and Wolf~\cite{iccv15_levy}          & 0.482 $\pm$ 0.615 & 0.45 &  0.065 $\pm$  0.092 & 0.90 & - & - \\
    Levy and Wolf$^*$~\cite{iccv15_levy}          & 0.237 $\pm$ 0.339 & 0.52 & 0.142 $\pm$ 0.231 & 0.73 & 0.286 $\pm$ 0.574 & 0.68 \\
    Runia~\etal~\cite{cvpr18_runia}       & 0.232 $\pm$ 0.344 & 0.62 & 0.103 $\pm$ 0.198 & 0.89 & - & - \\
    Runia~\etal~\cite{ijcv19_runia}       & 0.261 $\pm$ 0.396 & 0.62 & 0.094 $\pm$ 0.174 & 0.89 & - & - \\
    \hline
    Ours-Resnet18                       & 0.190 $\pm$ 0.327 & 0.70    & 0.062 $\pm$ 0.125 & 0.91    & 0.213 $\pm$ 0.343 & 0.69 \\
    Ours-Resnet50                       & 0.167 $\pm$ 0.293 & 0.75    & 0.081 $\pm$ 0.261 & 0.94   &  0.190 $\pm$ 0.288 & 0.74 \\
    Ours-Resnet101                      & \textbf{0.148 $\pm$ 0.290} & 0.75    & 0.066 $\pm$ 0.170 & 0.94    & 0.187 $\pm$ 0.303 & 0.77 \\
    Ours-Resnext101                     & 0.163 $\pm$ 0.311 & \textbf{0.76}    & \textbf{0.053 $\pm$ 0.115} & \textbf{0.95} & \textbf{0.147 $\pm$ 0.243} & \textbf{0.79} \\
    \hline
    \end{tabular}
    \end{center}
    \vspace{-6mm}\caption{Comparison with the existing methods on \emph{YTsegments}, \emph{QUVA Repetition} and \emph{UCFRep} for temporal repetition counting. The method with $^*$ is the re-implementation version by us trained on our \emph{UCFRep} benchmark.}\label{table:sota}\vspace{-3mm}
\end{table*}

\section{Experiments}
\label{sec:experiments}
\paragraph{Implementation Details.}
We implement the proposed network using Pytorch, and test it with an NVIDIA Geforce GTX1080Ti GPU. All input video frames of the network are resized to 112$\times$112, and we construct a $2M=32$ frames sequences. For training, we use Adam optimizer~\cite{kingma2014adam} with a fixed learning rate of 0.00005 and batch size of 24. We train our network on the \emph{UCFRep} with 100 epochs.

We train our network with the same pipeline of the proposed coarse-to-fine refinement.
Data augmentation is used to extend the annotations: if the variation of two consecutive repetitions is less than 0.3, we assume they are periodic. Then we add annotations within the interval automatically by linear interpolation.

During testing, we perform the coarse-to-fine refinement with $K=5$ stages. Our initial exhaustive searching is performed with 30 scales (ranging from 4 to $N/2$), and conduct 4 times refinement in the 1st and 2nd stages, 2 times in the 3rd stage, and 1 time in the 4th to 5th stages, leading to $30+4\cdot (1+2)+2\cdot 4+8+16=74$ forwards of the estimation network. The running time of our method depends on the times of the network forwards, and it takes averagely 1.8 seconds to process a video.

\paragraph{Evaluation Datasets.}
We evaluate our method on the three video datasets: the existing datasets \emph{YTsegments}~\cite{iccv15_levy} and \emph{QUVA Repetition}~\cite{cvpr18_runia}, as well as the proposed benchmark \emph{UCFRep}.
Both the \emph{YTsegments} and \emph{QUVA Repetition} contain 100 videos with a wide range of repetitions, like sports of humans and animal behaviors.
We consider all the videos from \emph{YTsegments} dataset and \emph{QUVA Repetition} dataset as testing set, and all the training and the validation is done on the proposed \emph{UCFRep} benchmark.
As a result, we split the videos in \emph{UCFRep} benchmark into the training set and validation set according to the group number from \emph{UCF101}. 421 videos with group numbers 1-20 are split into the training set, and 105 videos with group numbers 21-25 are in the validation set.

\paragraph{Evaluation Metric.}
Following the previous works~\cite{ijcv19_runia,iccv15_levy}, we evaluate the proposed method by counting accuracy.
For each dataset, we report the mean absolute error (MAE) and off-by-one-accuracy (OBOA) given $K$ videos
\begin{equation}
 {\rm MAE} = \frac{1}{K} \sum_{i=1}^{K}\frac{\left | \tilde{c}_i- c_i \right |}{\tilde{c}_i},
\end{equation}
\begin{equation}
 {\rm OBOA} = \frac{1}{K} \sum_{i=1}^{K}\left [{\left | \tilde{c}_i - c_i \right |}\leq 1 \right ],
\end{equation}
where $\tilde{c}$ is the ground truth repetition counts.
The mean absolute error is a widely used metric to directly evaluate counting errors. The off-by-one-accuracy can counts the rounding error and show the possible cycle cut-offs at both ends of the video as introduced in~\cite{ijcv19_runia}.

\begin{figure}[t]
    \begin{center}
        \includegraphics[width=0.8\linewidth]{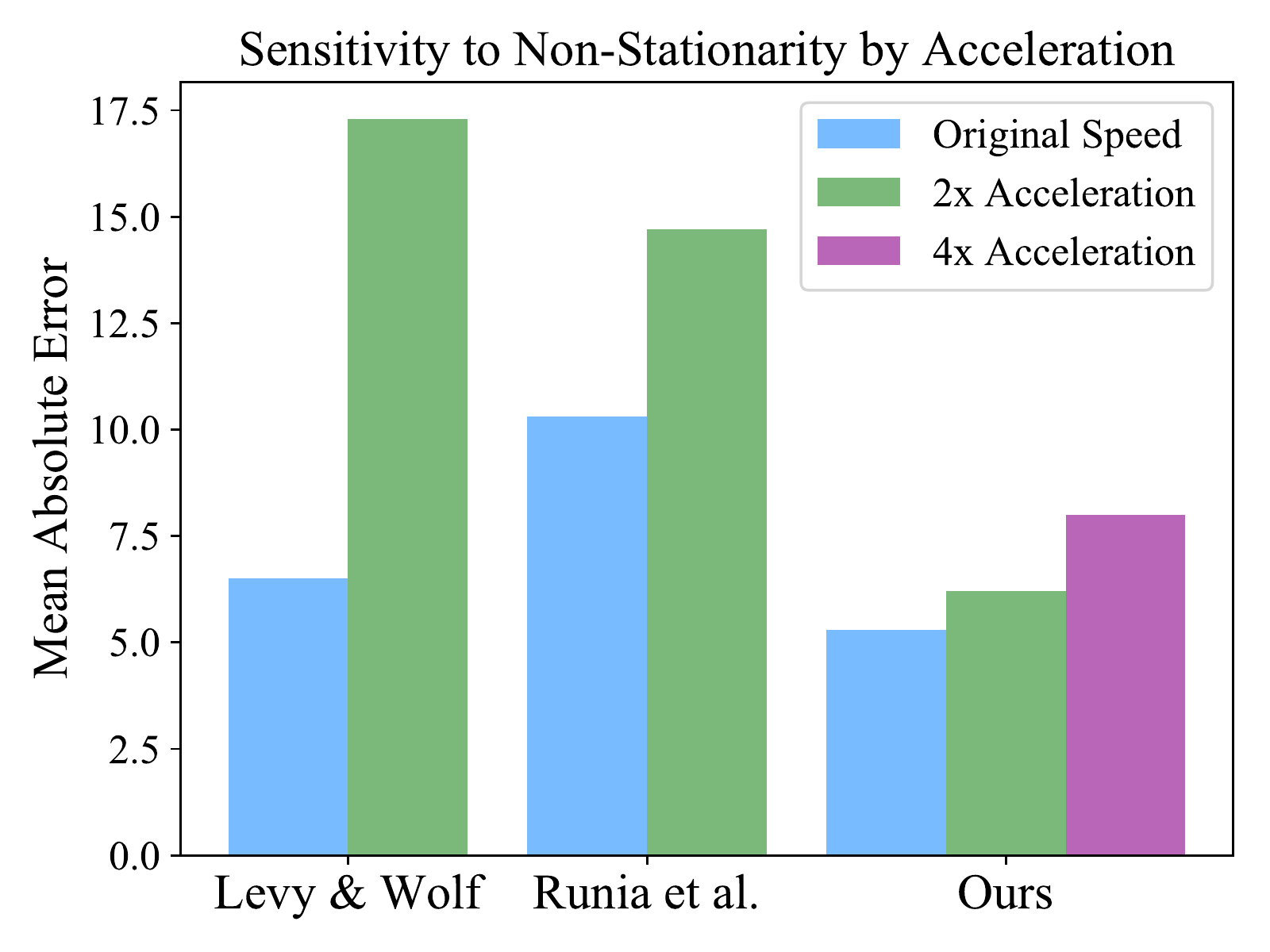}
    \end{center}
    \vspace{-6mm}\caption{Evaluation on the robustness to acceleration 1x, 2x and 4x on the \emph{YTsegments} dataset following the experiments in~\cite{cvpr18_runia}. Compared to a previous scale-insensitive method~\cite{cvpr18_runia}, our method is more robust to the time-scale. }\label{fig:speed}\vspace{-2mm}
\end{figure}

\begin{figure*}[t]
    \begin{center}
        \includegraphics[width=1.0\textwidth]{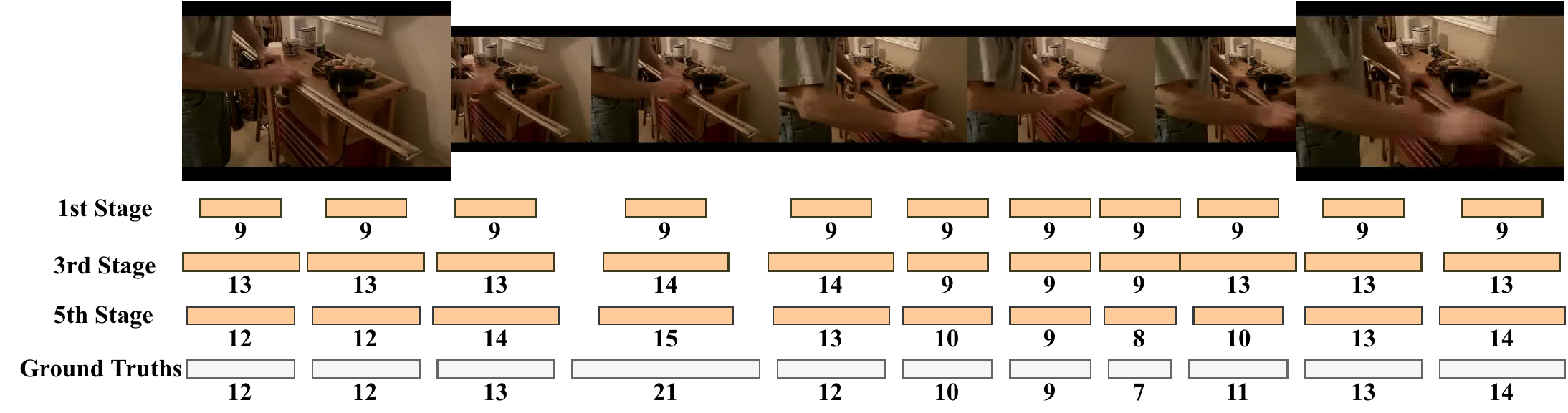}
    \end{center}
    \vspace{-6mm}\caption{Multi-stage cycle length visualization of a video from the \emph{QUVA Repetition} dataset. In this video, a man is painting (unseen during training), our coarse-to-fine strategy can progressively refine the cycle lengths.}\label{fig:visualization}\vspace{-3mm}
\end{figure*}

\subsection{Comparison with Other Methods}
The comparison with the existing methods for temporal repetition counting is shown in Table~\ref{table:sota}.
We compare our method with two hand-crafted feature methods~\cite{cvpr08_pogalin,ijcv19_runia} and one deep learning-based method~\cite{iccv15_levy}. As the complete source codes of~\cite{cvpr08_pogalin,ijcv19_runia} are unavailable, we compare to them on two previous testing datasets \emph{QUVA Repetition} and \emph{YTsegments}. We can observe that our method can outperform all the previous methods. It demonstrates that our method trained on the \emph{UCFRep} is general enough to the common repetitions from other datasets. Especially for the non-stationary dataset \emph{QUVA Repetition}, our method obtains improvement on MAE with 6.9$\%$ and OBOA with 14$\%$, indicating that our scale-insensitive framework can better handle the videos with varied cycle-length.

To demonstrate these improvements are brought mainly by the proposed framework rather than the new dataset, we fine-tune the learning-based method~\cite{iccv15_levy} on the new benchmark using our train/validation protocol. Note that the other two competitors~\cite{cvpr08_pogalin,ijcv19_runia} are training-free methods.
The original implementation~\cite{iccv15_levy} uses a simple 3D network to learn on synthesis data with 20 50$\times$50 images as input. We replace their network with Resnext101 to extract information from 32 112$\times$112 frames for adapting to the higher-dimensional data. We remove their ROI detection to keep the inference sequence similar to the training data, and the other implementations follow the published official code. Not surprisingly, because of the increased number of training data, the re-trained model on \emph{UCFRep} benchmark shows better performance compared with the original implementation on the \emph{QUVA Repetition} dataset. However, it cannot perform well on the periodic dataset \emph{YTsegments}, this is because their synthesis data is created following the restrict periodic assumption, while our dataset shows various types of repetitions. Compared with both the finetuned and original versions, our method outperforms them on all the datasets, as their network is designed to consider only a fixed scale of action. These results also demonstrate the success of our tailored context-aware and scale-insensitive framework.

We further evaluate the robustness to time-scale of our method. We follow~\cite{cvpr18_runia} to manually speed up the video to achieve different time-scales. As shown in Figure~\ref{fig:speed}, when the video is processed with different speeds, it poses a challenge to the fixed time-scale method~\cite{iccv15_levy} ($6.5\%$ on 1x and $17.3\%$ on 2x). Compared with the results ($10.3\%$ on 1x and $14.7\%$ on 2x) from the existing scale-insensitive method~\cite{cvpr18_runia}, our method is more robust to speed variations ($5.3\%$ on 1x, $6.2\%$ on 2x and $8.0\%$ on 4x), which implies that our method can detect the repetitions with different time-scales.

\begin{table}
    \begin{center}
    \begin{tabular}{lccc}
    \hline
    & MAE$\downarrow$ & OBOA$\uparrow$ & Iterations\\
    \hline\hline
    Stage 3   & 0.157 $\pm$ 0.284 & 0.78  & 50 \\ 
    Stage 4   & 0.156 $\pm$ 0.254 & 0.78  & 58 \\  
    Stage 5   & \textbf{0.147 $\pm$ 0.243} & \textbf{0.79}  & 74 \\  
    Stage 6   & 0.151 $\pm$ 0.254 & \textbf{0.79} & 106 \\  
    \hline
    \end{tabular}
    \end{center}
    \vspace{-6mm}\caption{Ablation study of the proposed coarse-to-fine refinement method on the \emph{UCFRep} benchmark validation set. }\label{table:ab_level}\vspace{-2mm}
\end{table}
\begin{table}
    \begin{center}
    \begin{tabular}{lcc}
    \hline
     &  MAE$\downarrow$ & OBOA$\uparrow$ \\
    \hline\hline
    Fixed   & 0.177 $\pm$ 0.280 & 0.70 \\
    Fixed+mAnchor & 0.171 $\pm$ 0.249 & 0.71 \\
    Free & 0.157 $\pm$ 0.243 & 0.76 \\
    Free+mAnchor & \textbf{0.147 $\pm$ 0.243} & \textbf{0.79} \\
    \hline
    \end{tabular}
    \end{center}
    \vspace{-6mm}\caption{Ablation study of the proposed context-aware estimation network on the \emph{UCFRep} benchmark validation set.}\label{table:ab_network}\vspace{-2mm}
\end{table}
\begin{table}
    \renewcommand\tabcolsep{2.5pt}
    \begin{center}
    \begin{tabular}{lccccc}
    \hline
    {Metric} & {All} & {HulaHoop}  & {Biking} & {Hammering}  & {Soccer}  \\
    \hline\hline
    MAE-avg  &0.147 &0.120 &0.123 &0.154 &0.168 \\
    MAE-std  &0.243 &0.240 &0.062 &0.170 &0.111 \\
    \hline
    \end{tabular}
    \end{center}
    \vspace{-6mm}\caption{Performance variations with respect to different action classes on the \emph{UCFRep} benchmark validation set.}\label{table:ab_var}
\end{table}

\subsection{Ablation Study}
We conduct the ablation study on \emph{UCFRep} validation set.
In Table~\ref{table:ab_level}, we compare the performance of our system utilizing different stages as the final stage in the coarse-to-fine refinement. The process with 6 stages will involve 32 iterations in the final stage, thus it overall needs $74+32=106$ iterations.
The results in this table indicate that involving more stages and computations in the refinement process can improve the results. We balance the trade-off between accuracy and speed, and choose stage 5 as the final stage.

We also compare the performance of our context-aware network with the other network designs in Table~\ref{table:ab_network}. We first compare the performance of using double time-scales for the two consecutive repetitions (Free) or single time-scale shared by the consecutive repetitions (Fixed).
The results with double time-scales are better than those with a single time-scale, which demonstrates that the free time-scales help to tackle the diverse cycle length.
In addition, the multi-anchors design (mAnchor) achieves the best performance integrated with the double time-scales. This implies that the regression can refine the cycle length with a large range, and thus benefitted from the multi-anchors prediction focusing on the diverse time-scales.

In Table~\ref{table:ab_var}, we further show the performance variations with respect to different action classes. We can see that the variations within the same action class are relatively small, indicating that our model is instance and class insensitive.

\subsection{Refinement Results Visualization}
To show the process of coarse-to-fine refinement, we visualize the prediction of the 1st stage, 3rd stage and the 5th stage over a video from \emph{QUVA Repetition} dataset in Figure~\ref{fig:visualization}. We set the each repetition prediction equal to the rounded mean value of the cycle length from the closet sampled position. From the results, we can find that we give an identical estimation to all the positions in stage 1 since it only involves one local prediction.
In the 3rd stage and 5th stage, the predictions after propagation and refinement achieve high overlap with the ground truth, showing that the proposed coarse-to-fine refinement can overcome the variation between consecutive repetitions. 

\section{Conclusion}
\label{sec::conclusion}
In this paper, we present a novel context-aware and scale-insensitive framework for temporal repetition counting. To tackle the challenges posed by the diverse cycle-lengths between videos and within repetitions, we propose a coarse-to-fine cycle refinement scheme. Instead of detecting the repetition with fixed time-scales, we search the time-scale with a wide range locally at the beginning and refine the scales for each temporal location in a coarse-to-fine manner. We further propose a context-aware regression network to learn contextual features for recognizing previous and future repetitions. The proposed network is designed to extract the context-aware features from two consecutive repetitions, and a anchor-based backend is tailored for detecting double-error or half-error. The proposed temporal repetition counting framework is evaluated and compared with state-of-the-art methods and achieves better results in the existing benchmarks as well as our newly proposed dataset.

\section*{Acknowledgement}
The work is supported by NSFC (Grant No. 61772206, U1611461, 61472145, 61702194, 61972162), Guangdong R$\&$D key project of China (Grant No. 2018B010107003, 2020B010165004, 2020B010166003), Guangdong High-level personnel program (Grant No. 2016TQ03X319), Guangdong NSF (Grant No. 2017A030311027), Guangzhou Key Project in Industrial Technology (Grant No. 201802010027, 201802010036), and the CCF-Tencent Open Research fund (CCF-Tencent RAGR20190112).

{\small
\bibliographystyle{ieee_fullname}
\bibliography{egbib}
}

\end{document}